\pdfoutput=1

\documentclass[11pt]{article}

\usepackage[]{EMNLP2022}

\usepackage{times}
\usepackage{latexsym}

\usepackage[T1]{fontenc}

\usepackage[utf8]{inputenc}

\usepackage{microtype}

\usepackage{inconsolata}
\usepackage{CJKutf8}
\usepackage{soul}
\usepackage{color, xcolor}

\usepackage{graphicx}
\usepackage{amsmath}
\usepackage{amssymb}
\usepackage{multirow}
\usepackage{makecell}
\usepackage[ruled,linesnumbered]{algorithm2e}
\SetKwInput{KwInput}{Input}                
\SetKwInput{KwOutput}{Output}              

\title{Leveraging Key Information Modeling to Improve Less-Data Constrained News Headline Generation via Duality Fine-Tuning}

\author{
    Zhuoxuan Jiang$^\dag$, Lingfeng Qiao$^\dag$, Di Yin$^\dag$, Shanshan Feng$^\ddag$, Bo Ren$^\S$ \\ $^\dag$Tencent Youtu Lab, Shanghai, China \\ $^\ddag$Harbin Institute of Technology, Shenzhen, China \\ $^\S$Tencent Youtu Lab, Hefei, China \\ \texttt{jzhx@pku.edu.cn}, \texttt{\{leafqiao,endymecyyin,timren\}@tencent.com}, \texttt{victor\_fengss@foxmail.com}
  }

\begin{document}
\maketitle
\begin{abstract}

Recent language generative models are mostly trained on large-scale datasets, while in some real scenarios, the training datasets are often expensive to obtain and would be small-scale. In this paper we investigate the challenging task of less-data constrained generation, especially when the generated news headlines are short yet expected by readers to keep readable and informative simultaneously. We highlight the key information modeling task and propose a novel duality fine-tuning method by formally defining the probabilistic duality constraints between key information prediction and headline generation tasks. The proposed method can capture more information from limited data, build connections between separate tasks, and is suitable for less-data constrained generation tasks. Furthermore, the method can leverage various pre-trained generative regimes, e.g., autoregressive and encoder-decoder models. We conduct extensive experiments to demonstrate that our method is effective and efficient to achieve improved performance in terms of language modeling metric and informativeness correctness metric on two public datasets.
\end{abstract}

\section{Introduction}

In an age of information explosion, headline generation becomes one fundamental application in the natural language process (NLP) field~\cite{fromstoh,survey}. Currently, the headline generation is usually regarded as a special case of general text summarization. Therefore, many cutting-edge techniques based on pre-trained models and fine-tuning methods can be directly adapted by feeding headline generation datasets~\cite{structure,representative}. Actually, compared with those textual summaries, headline generation aims at generating only one sentence or a piece of short texts given a long document (e.g., a news article). It is challenging to guarantee the generated headline readable and informative at the same time, which is important to attract or inform readers especially for news domain~\cite{truthful}.

Recently, some works find that neglecting the key information would degrade the performance of generative models which only consider capturing natural language~\cite{nan-etal-2021-improving}. Then many works about modeling different kinds of key information have been studied to enhance the information correctness of generative summaries. For example, overlapping salient words between source document and target summary~\cite{keyword_jd}, keywords~\cite{keyinformation}, key phrases~\cite{keyphrase} and named entities~\cite{entity} are involved to design generative models. However, those works are mostly either trained on large-scale datasets or targets for long summaries~\cite{pens}. In some real applications, it is expensive to obtain massive labeled data. Thus it becomes a much more challenging task that how to generate short headlines which should be both readable and informative under less-data constrained situations.

\begin{figure}[t]
\centering
\includegraphics[width=0.9\columnwidth]{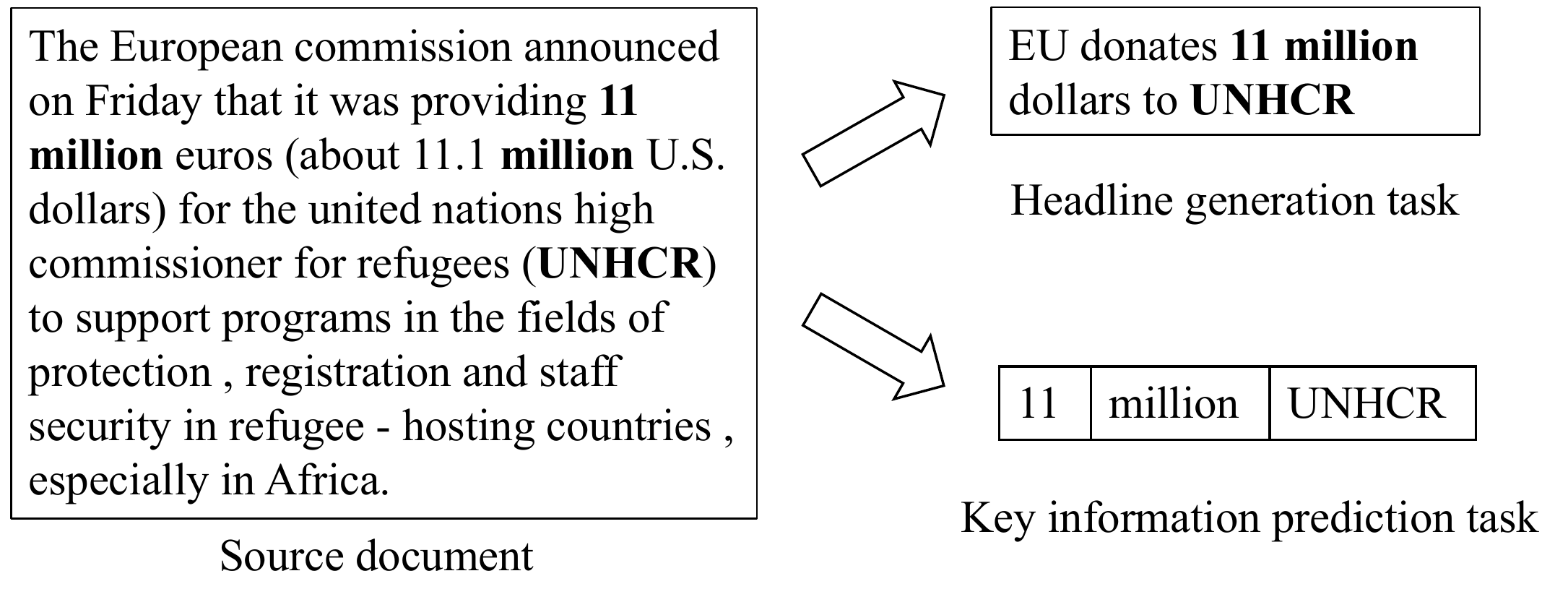}
\vspace{-2ex}
\caption{An example of multi-task decomposition for headline generation. The bold parts are salient tokens.}
\label{example}
\vspace{-3ex}
\end{figure}

To model the key information, existing works often follow the assumption that a generated summary essentially consists of two-fold elements: the natural language part and the key information part. The former focuses on language fluency and readability, while the later is for information correctness. For this reason, an additional task of key information prediction is leveraged and the multi-task learning method is employed~\cite{keyword_jd,entity}. Figure~\ref{example} can illustrate the intuitive idea more clearly, and the bold parts can be treated as the key information (overlapping salient tokens), which should be modeled well to inform correct and sufficient information for readers. 

To achieve the above motivation, technically, applying existing fine-tuning and multi-task learning methods to headline generation can be a natural choice. However they have some drawbacks. Firstly, single-task normal fine-tuning methods cannot explicitly model the key information well and hence reduce the informative correctness of generated headlines. Secondly, multi-task fine-tuning methods should improve the model ability by sharing the encoder and tailing two classifiers for key information prediction task and headline generation task, respectively. In fact, due to the limited dataset scale, the shared encoder could not be trained well to significantly distinguish the tasks or enhance each other mutually. As a result, vanilla multi-task methods could achieve little benefit for generation tasks~\cite{entity,magooda2021multitask}. Our empirical experiments later can also show this point. Therefore, existing single-task or multi-task fine-tuning methods cannot perform well under less-data constrained situations.

In this paper, we set out to address the above mentioned issues from the following two aspects. On the one hand, to explicitly model the key information, we still adopt the multi-task paradigm, while the two tasks utilize their own models. Then we argue that the two tasks have probabilistic connections and present them in dual forms. In this way, the key information is explicitly highlighted, and setting two separate models to obey duality constraints cannot only make the model more capable to distinguish tasks but also capture the relation between tasks. On the other hand, to capture more data knowledge from limited dataset, besides the source document, headlines and key tokens are additionally used as input data for the key information prediction task and headline generation task respectively. We call this method as \textbf{duality fine-tuning} which obeys the definition of dual learning~\cite{duallearning,duallearning2}. Moreover, we develop the duality fine-tuning method to be compatible with both autoregressive and encoder-decoder models (LM).

To evaluate our method, we collect two datasets with the key information of overlapping salient tokens\footnote{We expect our method to be orthogonal to specific key \par information definition.} in two languages (English and Chinese), and leverage various representative pre-trained models (BERT~\cite{bert}, UniLM~\cite{unilm} and BART~\cite{bart}). The extensive experiments significantly demonstrate the effectiveness of our proposed method to produce more readable (on Rouge metric) and more informative (on key information correctness metric) headlines than counterpart methods, which indicates that our method is consistently useful with various pre-trained models and generative regimes.

In summary, the main contributions include:

\begin{itemize}
\item We study a new task that how to improve performance of headline generation under less-data constrained situations. We highlight to model the key information and propose a novel duality fine-tuning method. To our best knowledge, this is the first work to integrate dual learning with fine-tuning paradigm for the task of headline generation. 

\item The duality fine-tuning method which should model multiple tasks to obey the probabilistic duality constraints is a new choice suitable for less-data constrained multi-task generation, in terms of capturing more data knowledge, learning more powerful models to simultaneously distinguish and build connections between multiple tasks, and being compatible with both autoregressive and encoder-decoder generative pre-trained models.

\item We collect two small-scale public datasets in two languages. Extensive experiments prove the effectiveness of our method to improve performance of readability and informativeness on Rouge metric and key information accuracy metric.
\end{itemize}

\section{Related Work}

Usually, headline generation is regarded as a special task of general abstractive text summarization, and the majority of existing studies could be easily adapted to headline generation by feeding headline related datasets~\cite{truthful,transformer-hg}. For example, sequence-to-sequence based models are investigated for text summarization, which emphasizes on generating fluent and natural summaries~\cite{seq2seq,rnn_summarization,cnn,pointer}. In recent years, the large-scale transformer-based models~\cite{bert,unilm,bart} and the two-stage (pre-training and fine-tuning) learning paradigm~\cite{pretraining,finetune,pretrain1} have greatly promoted the performance of most NLP tasks. And headline generation can also benefit from those works.

Since the length of headlines is often short and almost `every word is precious', compared to general text summarization, modeling the key information is better worth of paying attention~\cite{keyword_jd,keyphrase,overlap,entity,zhu-etal-2021-enhancing}. However, to our knowledge, little work focuses on this problem for headline generation, especially under the less-data constrained situations, and mostly they focus on low-resource long text summarization~\cite{parida,bajaj,yu2021adaptsum}. 


Recent years witness the rapid development of transformers-based pre-trained models~\cite{transformers} and two kinds of regimes of natural language generation (NLG) are prevalent~\cite{prefix}. One is based on autoregressive language models which have a shared transformer encoder structure for encoding and decoding~\cite{bert,unilm,roberta}, while the other is based on the standard transformer framework which has two separate encoder-decoder structures~\cite{bart,PEGASUS}. Fine-tuning and multi-task learning on them to reuse the ability of pre-trained models are widely studied for various tasks~\cite{pretrain2,pretrain1,pretrain3}. Our work can also align with this research line and we propose a new multi-task fine-tuning method.

We leverage the core idea of dual learning, which can fully mine information from limited data and well model multiple tasks by designing duality constraints~\cite{duallearning,duallearning2}. This learning paradigm has been successfully applied to many fields, such as image-to-image translation~\cite{image_trans}, recommendation system~\cite{dualrecommendation}, supervise and unsupervised NLU and NLG~\cite{su_nlunlg,su_dual2020}. Those works have demonstrated that duality modeling is suitable for small-scale training situations.

\begin{figure*}[t]
	\centering
	\includegraphics[width=2\columnwidth]{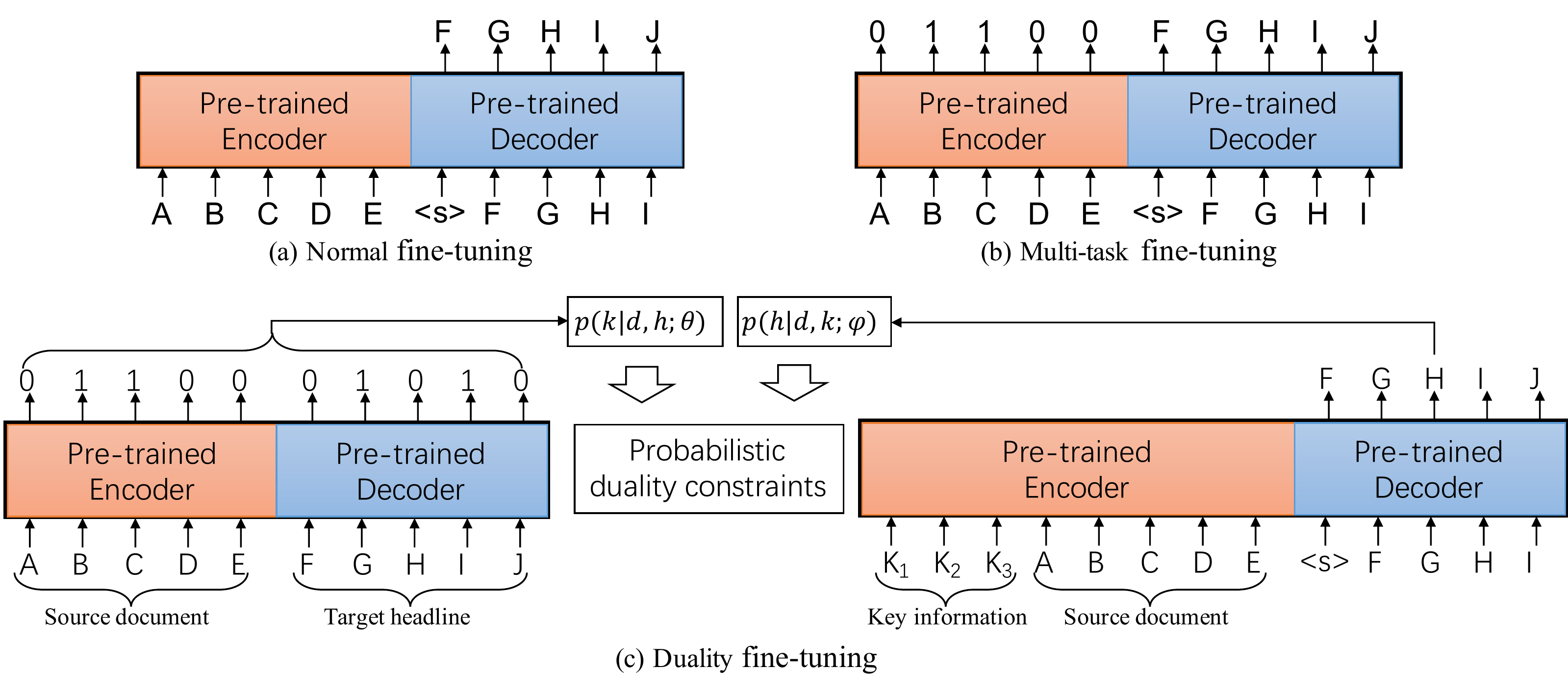}
	\caption{The overview of different fine-tuning methods. (a) is normal fine-tuning for single-task headline generation. (b) is multi-task fine-tuning which has an additional task of predicting the salient tokens among inputs with the encoder. (c) is the proposed duality fine-tuning which owns two separate models and more information as input by sticking to probabilistic duality constraints. Note that all the paired pre-trained encoder and decoder can be instanced as autoregressive LM (e.g., UniLM) or encoder-decoder (e.g., BART) regimes.}
	\label{dualkey}
\end{figure*}

\section{Problem Definition}

In this section, we formally present our problem. The training set is denoted as $\mathcal{X}=(\mathcal{D},\mathcal{H},\mathcal{K})$, where $\mathcal{D}$ and $\mathcal{H}$ are the sets of source documents and target headlines. $\mathcal{K}$ is the set of key information, which indicates the overlapping salient tokens (stopwords excluded) in each pair of document and headline. A training sample is denoted as a tuple $(d,h,k)$. $d=\{x_1^{(d)},x_2^{(d)},...,x_n^{(d)}\}$, $h=\{x_1^{(h)},x_2^{(h)},...,x_m^{(h)}\}$, $k=\{x_1^{(k)},x_2^{(k)},...,x_l^{(k)}\}$, where $x_i^{(*)}$ is a token of document, headline or key information, and $n$, $m$, $l$ are the lengths of respective token sequences.

\subsection{Definition of Dual Tasks}

Given the input data $x=(d,h,k)$, we define our problem in a dual form, which contains two tasks. Formally, the key information prediction task aims at finding a function $f: (d,h)\to k$, which maximizes the conditional probability $p(k|d,h;\theta)$ of the real key information $k$. Correspondingly, the headline generation task targets at learning a function $g: (d,k)\to h$, which maximizes the conditional probability $p(h|d,k;\varphi)$ of real headline $h$. The two tasks can be defined as follows:
\begin{equation}
\footnotesize
\begin{aligned}
    f(d,h;\theta)\triangleq\arg\max\prod_{x\in\mathcal{X}}p(k|d,h;\theta),\\
    g(d,k;\varphi)\triangleq\arg\max\prod_{x\in\mathcal{X}}p(h|d,k;\varphi).\nonumber
\end{aligned}
\end{equation}

\subsection{Probabilistic Duality Constraints}

Based on the principle of dual learning paradigm~\cite{duallearning}, we treat the key information prediction task as primary task and the headline generation task as secondary task. Ideally, if the primary model and secondary model are both trained optimally, the probabilistic duality between the two tasks should satisfy the following equation:
\begin{equation}
\footnotesize
\begin{aligned}
    p(\mathcal{X})=&\prod_{x\in\mathcal{X}}P(d,k,h) = \prod_{x\in\mathcal{X}}p(d)p(h|d;\hat{\varphi})p(k|d,h;\theta) \\
    = &\prod_{x\in\mathcal{X}}p(d)p(k|d;\hat{\theta})p(h|d,k;\varphi).\nonumber
\end{aligned}
\label{eq_constraint}
\end{equation}

$p(k|d,h;\theta)$ and $p(h|d,k;\varphi)$ are the target models to learn, while $p(k|d;\hat{\theta})$ and $p(h|d;\hat{\varphi})$ denote the marginal distribution models. By integrating the above probabilistic duality equation and further dividing the common term $p(d)$, our problem can be formally defined to optimize the objectives:
\begin{equation}
\footnotesize
\begin{aligned}
    &\text{Objective 1}:\min_\theta\frac{1}{|\mathcal{X}|}\sum_{x\in\mathcal{X}}l_1(f(d,h;\theta),k),\\
    &\text{Objective 2}:\min_\varphi\frac{1}{|\mathcal{X}|}\sum_{x\in\mathcal{X}}l_2(g(d,k;\varphi),h),\\
    \text{s.t.} \prod_{x\in\mathcal{X}}&p(h|d;\hat{\varphi})p(k|d,h;\theta)=\prod_{x\in\mathcal{X}}p(k|d;\hat{\theta})p(h|d,k;\varphi),
\end{aligned}
\label{eq_problem}
\end{equation}
where $l_1$ is the loss function for key information prediction and $l_2$ is that for headline generation.
\section{Duality Fine-tuning Methodology}

\subsection{Overview}

Before introducing the duality fine-tuning method, we would review the normal fine-tuning and multi-task fine-tuning methods. As shown in Figure~\ref{dualkey}, the (a) normal fine-tuning method is single-task and optimizes the generative model with new dataset by leveraging the same structure of pre-trained models. To explicitly model the key information, (b) multi-task fine-tuning method would use an additional task to binarily predict salient tokens, where 1 means key information and 0 means not. Here the two tasks share the common encoder. 

Different from the above two methods, although the (c) duality fine-tuning method is also a multi-task paradigm, however it shows totally different structure and process in terms of the following three aspects. Firstly, the two tasks own their respective encoder and decoder pairs inherited from a consistent pre-trained model structure. Secondly, the each model can be fed with more input information than normal and multi-task fine-tuning, i.e. key information prediction task can further utilize the headline data while headline generation task can extra utilize the data of key tokens. Thirdly, the two tasks should stick to the probabilistic duality constraints to build connections between the two tasks by Eq.~\ref{eq_problem}.

Note that all the three methods in Figure~\ref{dualkey} are compatible with autoregressive language models (the encoder and decoder are integrated in one transformer encoder like UniLM) and encoder-decoder models (standard transformer structure like BART).

\subsection{Model for Key Information Prediction}

Given the pair of source document and target headline as inputs, we expect the model to predict the key information and learn the pattern that the information is present at both sides. We regard the prediction task as binary classification for every token: $\hat{y}^{(k)}=p(k|d,h;\theta)=p(y^{(k)}|x^{(d)},x^{(h)};\theta)=\{0,1\}^{n+m}$. The last hidden state layers of encoder and decoder are tailed with the multi-layer perception (MLP) to make binary predictions by using sigmoid classifier.

If the relied pre-trained model is autoregressive, the encoder and decoder would belong to a shared transformer encoder structure, and if the encoder-decoder pre-trained model is leveraged, there can be a standard transformer structure. The objective function $l_1$ of Objective 1 in Eq.~\ref{eq_problem} can be rewritten by using the cross entropy loss function:
\begin{equation}
	\footnotesize
	l_1=-\sum_{z=1}^{n+m}(y_z^{(k)}\log(\hat{y}_z^{(k)})+(1-y_z^{(k)})\log(1-\hat{y}_z^{(k)})).
	\label{eq_loss_1}
\end{equation}

\subsection{Model for Headline Generation}

Given the source document and key information, we expect the model to learn that the tokens put ahead source document are explicitly highlighted and they are important to generate headlines. The generation process of headline is by once a token and generating current token is based on attending the key information, source document and already generated tokens. The formal calculation of predicting the $j$-th token is: $\hat{y}_j^{(h)}=p(y_j^{(h)}|x^{(d)},x^{(k)},y_{<j}^{(h)};\varphi)$. The last hidden state layer of the decoder is connected by a softmax function to generate tokens one by one. The details of generation process can be referred from the original literatures of adopted pre-trained models.

Similar to the corresponding key information prediction task, the same transformer encoder structure is adopted for autoregressive LMs and the standard transformer structure is for encoder-decoder LMs. The objective function $l_2$ of Objective 2 in Eq.~\ref{eq_problem} can be formally rewritten by using the cross entropy loss function:
\begin{equation}
	\footnotesize
	l_2=-\sum_{j=1}^{m}y_j^{(h)}\log(\hat{y}_j^{(h)}).
	\label{eq_loss_2}
\end{equation}


\subsection{Training \& Testing by Duality Fine-tuning}

To optimize the Objective 1 and Objective 2 under the duality constraints in Eq.~\ref{eq_problem}, we transform the constraint as a calculable regularization term:
\begin{equation}
\footnotesize
\begin{aligned}
    l_{duality}&=\sum_{x\in\mathcal{X}}[\log p(h|d;\hat{\varphi})+\log p(k|d,h;\theta)\\&-\log p(k|d;\hat{\theta})-\log p(h|d,k;\varphi)]^2,
\end{aligned}
\label{eq_duality_1}
\end{equation}

where $p(k|d;\hat{\theta})$ and $p(h|d;\hat{\varphi})$ are the marginal distribution models for key information prediction and headline generation respectively. 

\paragraph{Marginal Distribution Models} We define the marginal distribution models to calculate the duality regularization term $l_{duality}$. The marginal models can be obtained by just simplifying their corresponding dual models. For example, marginal key information prediction model is single-task token classification and only adopts the encoder part as $p(\mathcal{K}|\mathcal{D};\hat{\theta})=\prod_{x\in\mathcal{X}}\prod_{i=1}^{n}p(x_i^{(d)})$, while marginal headline generation is the normal fine-tuning task by calculating $p(\mathcal{H}|\mathcal{D};\hat{\varphi})=\prod_{x\in\mathcal{X}}\prod_{j=1}^{m}p(y_{j}^{(h)}|x^{(d)},y_{<j}^{(h)})$.

Since the two marginal distribution models are only involved in the calculation of regularization term $l_{duality}$ and will not be updated during the process of training dual models, they could be offline trained in advance. So in order to save the memory cost during duality fine-tuning, the predicted marginal key information, generated marginal headlines and their losses for each training sample can be calculated and stored beforehand.

\paragraph{Dual Model Training} After defining the duality regularization term and marginal models, we can obtain the calculable loss functions for duality fine-tuning by combining Eq.\ref{eq_problem} and Eq.\ref{eq_duality_1} as the following:

\vspace{-4ex}
\begin{equation}
\footnotesize
\begin{aligned}
    \mathcal{L}_1&=\min_\theta\frac{1}{|\mathcal{X}|}\sum_{x\in\mathcal{X}}(-\sum_{z=1}^{n+m}(y_z^{(k)}\log(\hat{y}_z^{(k)})\\&+(1-y_z^{(k)})\log(1-\hat{y}_z^{(k)}))+\lambda_1l_{duality}),
\end{aligned}
\label{eq_loss_11}
\end{equation}
\begin{equation}
\footnotesize
\begin{aligned}
    \mathcal{L}_2=\min_\theta\frac{1}{|\mathcal{X}|}\sum_{x\in\mathcal{X}}(-\sum_{j=1}^{m}y_j^{(h)}\log(\hat{y}_j^{(h)})+\lambda_2l_{duality}),
\end{aligned}
\label{eq_loss_22}
\end{equation}
where $\lambda_1$ and $\lambda_2$ denote the weights of the duality terms to control the impact of the duality constraints on the model optimization. The detailed algorithm for training is described in Algorithm~\ref{alg:algorithm_1}. Line 1-2 denote the model pre-training and parameter initialization. Line 5-12 are the one-step optimization for a mini-batch of training data, and the model should compute (or retrieve) the marginal losses and model losses ($l_1$ and $l_2$) successively. 

\begin{algorithm}
	\footnotesize
	\caption{\footnotesize{Training for Duality Fine-tuning}}
	\label{alg:algorithm_1}
	\KwInput{The training dataset $\mathcal{X}=[\mathcal{D},\mathcal{H},\mathcal{K}]$}
	\KwOutput{Dual model parameters $\theta$ and $\varphi$}
	Pre-train marginal models $p(k|d;\hat{\varphi})$ and $p(h|d;\hat{\theta})$\;
	Initialize all trainable parameters of $p(k|d,h;\theta)$ and $p(h|d,k;\varphi)$, set $t=1$\;
	\While{$t<T$}{
		\ForEach{mini-batch [$d$,$h$,$k$]}{
			Compute (or retrieve) marginal losses\;
			Compute model losses with Eq.\ref{eq_loss_1} and Eq.\ref{eq_loss_2}\;
			Update dual model losses by Eq.\ref{eq_loss_11} and Eq.\ref{eq_loss_22}\;
			Optimize $\theta$ for dual model $p(k|d,h;\theta)$\;
			Optimize $\varphi$ for dual model $p(h|d,k;\varphi)$\;
		}
	}
	\textbf{return} optimized $\theta$ and $\varphi$.
\end{algorithm}

\paragraph{Dual Model Testing} In the testing stage, we only have the documents as input and do not have the real key information and headlines. In order to save the run-time memory and computing resource cost, we use an open tool spaCy\footnote{https://spacy.io/} to extract the key information from the source document to approximate the tokens predicted by the dual key information prediction model, and therefore only one dual model, i.e., the dual headline generation model, is loaded into memory for making generation. 

\begin{table*}
	\centering\small
	\resizebox{\linewidth}{!}{
	\begin{tabular}{clccc||ccc|ccc}
		\hline
		\multirow{2}{*}{\makecell[c]{Pre-trained\\Model}} & \multirow{2}{*}{\makecell[c]{Fine-tune\\Method}} & \multicolumn{3}{c||}{} & \multicolumn{3}{c|}{micro} & \multicolumn{3}{c}{macro} \\ 
		& & Rouge-1 & Rouge-2 & Rouge-L & $\text{prec}_t$ & $\text{recall}_t$ & $\text{F1}_t$ & $\text{prec}_t$ & $\text{recall}_t$ & $\text{F1}_t$ \\
		\hline
		\multirow{4}{*}{\makecell[c]{BERT}} & Normal  & 0.3598 & 0.1626 & 0.3421 & 44.06 & 52.76 & 48.02 & 44.78 & \textbf{53.19} & 48.63 \\
		& Normal+ & 0.3594 & 0.1483 & 0.3411 & \textbf{56.94} & 46.15 & 50.98 & \textbf{58.67} & 49.08 & \textbf{53.45} \\
		& Multi-task & 0.3672 & \textbf{0.1775} & \textbf{0.3500} & 45.23 & \textbf{52.79} & 48.72 & 45.78 & 52.79 & 49.03 \\
		& Duality & \textbf{0.3692} & 0.1627 & 0.3469 & 51.20 & 51.36 & \textbf{51.28} & 51.50 & 51.44 & 51.47 \\
		\hline
		\multirow{4}{*}{\makecell[c]{UniLM}} & Normal & 0.3663 & 0.1739 & 0.3489 & 42.10 & 53.55 & 47.14 & 42.80 & 53.90 & 47.71 \\
		& Normal+ & 0.3524 & 0.1450 & 0.3285 & \textbf{53.57} & 48.49 & 50.90 & \textbf{54.43} & 51.57 & 52.96 \\
		& Multi-task & 0.3557 & 0.1631 & 0.3365 & 40.10 & 54.00 & 46.03 & 41.21 & 54.45 & 46.91 \\
		& Duality & \textbf{0.4025} & \textbf{0.1896} & \textbf{0.3774} & 45.12 & \textbf{60.88} & \textbf{51.82} & 47.50 & \textbf{61.09} & \textbf{53.45} \\
		\hline
		\multirow{4}{*}{\makecell[c]{BART}} & Normal & 0.4798 & 0.2753 & 0.4496 & 53.05 & 67.67 & 59.48 & 54.57 & 68.51 & 60.75 \\
		& Normal+ & 0.5005 & 0.2829 & 0.4711 & 56.71 & 70.24 & 62.75 & 58.72 & 70.67 & 64.14 \\
		& Multi-task & 0.4765 & 0.2699 & 0.4491 & 52.92 & 66.81 & 59.06 & 54.05 & 67.54 & 60.04 \\
		& Duality & \textbf{0.5372} & \textbf{0.3097} & \textbf{0.4999} & \textbf{62.12} & \textbf{79.57} & \textbf{69.77}  & \textbf{63.73} & \textbf{79.79} & \textbf{70.86} \\
		\hline
	\end{tabular}
	}
	\caption{\label{table:result_ggw} Comparison of Rouge and key information accuracy (\%) on Gigaword-3k dataset.}
\end{table*}

\begin{table*}
	\centering\small
	\resizebox{\linewidth}{!}{
	\begin{tabular}{clccc||ccc|ccc}
		\hline
		\multirow{2}{*}{\makecell[c]{Pre-trained\\Model}} & \multirow{2}{*}{\makecell[c]{Fine-tune\\Method}} & \multicolumn{3}{c||}{} & \multicolumn{3}{c|}{micro} & \multicolumn{3}{c}{macro} \\ 
		& & Rouge-1 & Rouge-2 & Rouge-L & $\text{prec}_t$ & $\text{recall}_t$ & $\text{F1}_t$ & $\text{prec}_t$ & $\text{recall}_t$ & $\text{F1}_t$ \\
		\hline
		\multirow{4}{*}{\makecell[c]{BERT}} & Normal & 0.4109 & 0.2722 & 0.3891 & 56.68 & 50.20 & 53.24 & 56.71 & 49.62 & 52.93 \\
		& Normal+ & 0.4164 & 0.2471 & 0.3893 & 71.85 & 45.93 & 56.04 & 72.45 & 45.76 & 56.09 \\
		& Multi-task & 0.4277 & 0.2835 & 0.4045 & 59.30 & 51.89 & 55.35 & 59.20 & 51.37 & 55.00 \\
		& Duality & \textbf{0.5279} & \textbf{0.3321} & \textbf{0.4807} & \textbf{73.64} & \textbf{59.68} & \textbf{65.93} & \textbf{74.24} & \textbf{59.53} & \textbf{66.07} \\
		\hline
		\multirow{4}{*}{\makecell[c]{UniLM}} & Normal  & 0.4137 & 0.2806 & 0.3905 & 56.37 & 51.06 & 53.58 & 55.98 & 50.16 & 52.91 \\
		& Normal+ & 0.4152 & 0.2502 & 0.3875 & 68.13 & 48.15 & 56.42  & 69.15 & 47.93 & 56.62 \\
		& Multi-task & 0.4147 & 0.2788 & 0.3909 & 52.68 & 53.51 & 53.09 & 53.28 & 52.54 & 52.91 \\
		& Duality & \textbf{0.5128} & \textbf{0.3324} & \textbf{0.4636} & \textbf{69.72} & \textbf{58.71} & \textbf{63.74} & \textbf{70.56} & \textbf{58.22} & \textbf{63.80} \\
		\hline
		\multirow{4}{*}{\makecell[c]{BART}} & Normal & 0.4301 & 0.2943 & 0.3992 & 49.68 & 56.93 & 53.06 & 50.62 & 56.02 & 53.18 \\
		& Normal+ & 0.5176 & 0.3338 & 0.4332 & 64.43 & 60.37 & 62.33 & 67.34 & 60.06 & 63.49 \\
		& Multi-task & 0.4239 & 0.2882 & 0.3937 & 49.76 & 55.81 & 52.61 & 50.73 & 54.96 & 52.76 \\
		& Duality & \textbf{0.6636} & \textbf{0.4720} & \textbf{0.5766} & \textbf{74.98} & \textbf{79.73} & \textbf{77.29} & \textbf{75.43} & \textbf{79.16} & \textbf{77.25} \\
		\hline
	\end{tabular}
	}
	\caption{\label{table:result_thu} Comparison of Rouge and key information accuracy (\%) on THUCNews-3k dataset.}
\end{table*}

\section{Experiments}
\subsection{Datasets}

To evaluate the duality fine-tuning's effectiveness, we collect two public corpora, Gigaword~\cite{ggw} and THUCNews~\cite{thucnews}. The overlapping words (stop-words excluded) between each pair of source document and target headline are regarded as the key information. 

\textbf{Gigaword} is in English and collected from news domain. We randomly extract 3,000/500/500 samples for model training/validating/testing from the original corpus\footnote{https://github.com/harvardnlp/sent-summary}, to approximate a less-data constrained situation. Here all the samples must contain key information. 

\textbf{THUCNews} is in Chinese and collected from the Sina News website\footnote{http://thuctc.thunlp.org/}. Each sample contains a headline and a news article.
We pre-process this dataset by also randomly extracting 3,000/500/500 training/validating/testing samples and all of them contain key information.

\subsection{Baselines and Metrics}


We compare the duality fine-tuning (Duality) with normal fine-tuning (Normal) and multi-task fine-tuning methods (Multi-task). Additionally, the Normal method has a variant (Normal+) that replaces the original input (source document) with key-token-enhanced input (key tokens+source document). We adopt base-scale versions of BERT, UniLM and BART as pre-trained models which are all representative either for autoregressive LMs or encoder-decoder regimes among NLG tasks.

We use the F1-version Rouge~\cite{rouge} to measure the comprehensive performance of language modeling on both the token-level precision and recall factors. To evaluate the informativeness accuracy, macro and micro $\text{prec}_t$, $\text{recall}_t$, and $\text{F1}_t$~\cite{entity} (denoting precision, recall, and F1 between generated and ground-truth salient tokens) are used. Readers can refer to the literature for details of calculating formulas.

\begin{figure*}[t]
	\centering
	\includegraphics[width=2\columnwidth]{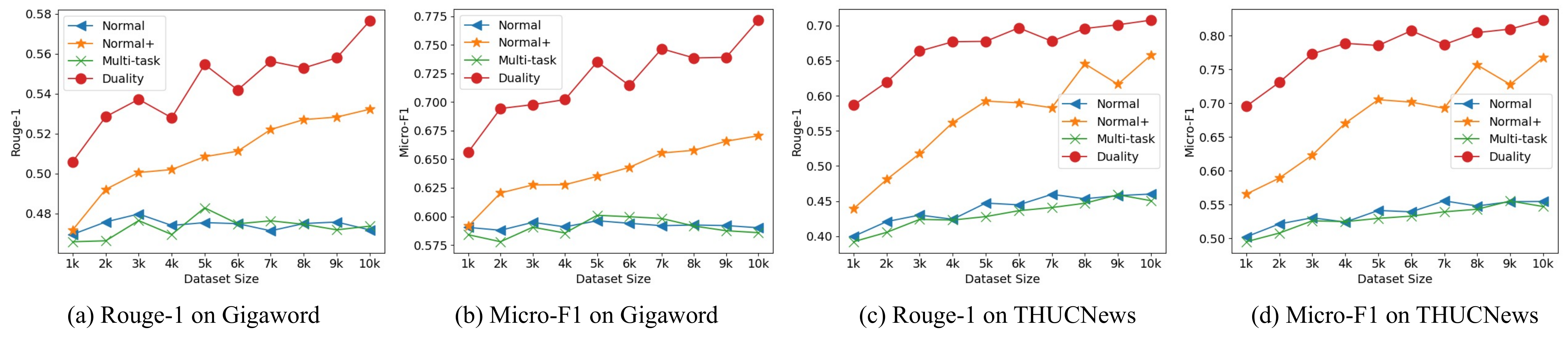}
	\caption{Performance of Rouge-1 and Micro-F1 on different sizes of THUCNews and Gigaword training datasets.}
	\label{datasize}
\end{figure*}

\subsection{Experimental Settings}

In all experiments, we keep the consistent default parameters with the pre-trained models during fine-tuning. All the models are trained for at least 10 epochs, and the experimental results are the average values from 5 runs of modeling learning. 
The batch size is set as 64 for normal/multi-task/marginal training and 16 for duality training, since dual learning would occupy more memory to reflect two models. However, during validating and testing phases, all the methods would spend the similar memory and computing resources. The learning rate is set 1e-5 for English dataset and 5e-5 for Chinese dataset. The max lengths of document and headline tokens for Gigaword is set 192 and 64, and those for THUCNews are 512 and 30. The beam search size for testing is set 5. Empirically by trying a grid search strategy, we set $\lambda_1=0.2$, $\lambda_2=0.8$ to emphasize the dual task of headline generation. Other detailed parameters can refer to the original literature of pre-trained models.

\subsection{Automatic Evaluation}

\begin{table}
	\centering\small
	\begin{tabular}{lcccc}
		\hline
		\multicolumn{1}{c}{\multirow{2}{*}{Method}} & \multicolumn{2}{c}{Gigaword}    & \multicolumn{2}{c}{THUCNews}     \\ 
		\cline{2-5} \multicolumn{1}{c}{} & Read. & Info. & Read. & Info. \\
		\hline
		Reference & 4.40 & 4.29 & 4.79 & 4.78 \\
		\hline
		Normal & 3.75 & 3.44 & 3.41 & 3.06 \\
		Multi-task & 3.67 & 3.58 & \textbf{3.97} & 3.29 \\
		Duality & \textbf{3.77} & \textbf{4.00} & 3.90 & \textbf{3.51} \\
		\hline
	\end{tabular}
\caption{\label{human}Human evaluation on readability (Read.) and informativeness (Info.) of generated headlines.}
\end{table}
\paragraph{Performance on 3K datasets} We adopt the data size of 3,000 (3K) to approximate the less-data constrained situation, because usually it is easy to hand-crafted label 3K (or comparable quantity) samples. Table~\ref{table:result_ggw} and Table~\ref{table:result_thu} present the performance of generation (left part) and key information accuracy (right part) on Gigaword-3k dataset and THUCNews-3k dataset, respectively. From the left part in Table~\ref{table:result_ggw}, we find Duality fine-tuning method can achieve the superior scores almost with all the pre-trained models. From the right part for key information accuracy (micro and macro $\text{prec}_t$, $\text{recall}_t$ and $\text{F1}_t$ ), duality fine-tuning method can also greatly enhance the informative correctness, especially using BART as pre-trained models.

From the left part of Table~\ref{table:result_thu}, Duality fine-tuning method performs much better than Normal (and Normal+) fine-tuning and Multi-task fine-tuning methods. The table's right part also suggests the consistent effectiveness that duality method can generate more informative and accurate headlines with small-scale training datasets. Comparing with Table~\ref{table:result_ggw} and Table~\ref{table:result_thu}, the results may indicate that duality fine-tuning should be more suitable for Chinese than English datasets due to the more stable and higher observed improvement with different pre-trained models.

The two tables could reflect some observations. First, our duality fine-tuning method is generally and effectively applied to various generative pre-trained models, e.g. autoregressive LM (BERT and UniLM) and encoder-decoder (BART) regimes. Then, our method performs much better on BART than on the others, we think, because encoder-decoder models have separate transformer networks instead of only adopting the encoder structure, providing the more powerful model ability and larger model scale, which is friendly for less-data constrained situations. Moreover, the results in the two tables can also demonstrate that Duality fine-tuning method is effective to capture more data knowledge from limited data by using two separate dual models corresponding to tasks, and the designed probabilistic duality constraints are effective to build connections and enhance generation.

\sethlcolor{lightgray}
\begin{CJK}{UTF8}{gbsn}
\begin{table*}[ht]
	\centering\small
	\resizebox{\linewidth}{!}{
	\begin{tabular}{|l|l|l|l|}
		\hline
		\multicolumn{4}{|c|}{Cases from the Gigaword dataset} \\
		\hline
		Ground Truth & Normal & Multi-task & Duality \\
		\hline
		\makecell[l]{\hl{german union} urges \\ members to down \\ tools friday in \hl{iraq} protest} & 
		\makecell[l]{\hl{german} industrial \\ \hl{union} urges workers \\ to stop work} &
		\makecell[l]{\hl{german}'s largest \\ industrial \hl{union} urges \\ workers to stop work} & 
		\makecell[l]{\hl{german} labor \hl{union} \\ urges workers to stop \\ work over \hl{iraq} war}
		\\
		\hline
		\makecell[l]{\hl{bourdais} beats \hl{tracy} \\in champ car opener} &
		\makecell[l]{newman - haas wins \\ toyota grand prix} &
		\makecell[l]{newman - haas wins \\ toyota grand prix} &
		\makecell[l]{\hl{bourdais} beats \hl{tracy} \\ to win toyota grand prix} \\
		\hline
		\makecell[l]{\hl{iran} deal to ship \\ \hl{uranium} abroad meets \\ skepticism} &
		\makecell[l]{\hl{iran} says it will \\ continue enrichment} &
		\makecell[l]{\hl{iran} agrees to big \\ power demand} &
		\makecell[l]{\hl{iran} agrees to nuclear \\ enrichment but insists \\ it will continue enrich \hl{uranium}} \\
		\hline
		\makecell[l]{\hl{rockets} fired at suspected \\ \hl{us base} in \hl{pakistan}} &
		\makecell[l]{\hl{rockets} fired at \\ \hl{pakistan base}} &
		\makecell[l]{\hl{rockets} fired at \\ northwest \hl{pakistan base}} &
		\makecell[l]{\hl{rockets} fired at \hl{us base} \\ in \hl{pakistan}} \\
		\hline
		\makecell[l]{\hl{israeli} army destroys \\ \hl{palestinian} homes in \hl{rafah}} &
		\makecell[l]{\hl{israeli} troops operate \\ in \hl{rafah}} &
		\makecell[l]{\hl{israeli} army tanks  \\ operate in \hl{rafah}} &
		\makecell[l]{\hl{israel} starts operation in \\ \hl{palestinian} - controlled \hl{rafah}} \\
		\hline
		\hline
		\multicolumn{4}{|c|}{Cases from the THUCNews dataset} \\
		\hline
		Ground Truth & Normal & Multi-task & Duality \\
		\hline
		\makecell[l]{\colorbox{lightgray}{at\&t业绩}未受\colorbox{lightgray}{verizon推iphone}\\明显影响} & 
		\makecell[l]{\colorbox{lightgray}{at\&t}第一季度新增160万\\非手机联网设备} &
		\makecell[l]{\colorbox{lightgray}{at\&t}第一季度新增160万\\非手机联网设备} & 
		\makecell[l]{\colorbox{lightgray}{at\&t}第一季度\colorbox{lightgray}{业绩}没有受\\到\colorbox{lightgray}{verizon推}出\colorbox{lightgray}{iphone}影响} \\
		\makecell[l]{Translation: \hl{at\&t's performance}\\ is not significantly affected by \\\hl{Verizon's launch of iPhone}} & 
		\makecell[l]{Translation: \hl{at\&t} added \\1.6 million non-mobile \\internet-connected devices\\ in the first quarter} &
		\makecell[l]{Translation: \hl{at\&t} added \\1.6 million non-mobile \\internet-connected devices\\ in the first quarter} & 
		\makecell[l]{Translation: \hl{at\&t}'s first-\\quarter \hl{performance} were not\\ affected by \hl{Verizon's launch}\\\hl{of the iPhone}} \\
		\hline
		\makecell[l]{\colorbox{lightgray}{2gb内存320gb硬盘联想}\\\colorbox{lightgray}{b460el}仅\colorbox{lightgray}{2699元}} & 
		\makecell[l]{gt芯t3500芯\colorbox{lightgray}{联想b460el}\\-tth仅售\colorbox{lightgray}{2699元}} &
		\makecell[l]{i3芯t3500芯\colorbox{lightgray}{联想b460el}\\-tth仅售\colorbox{lightgray}{2699元}} & 
		\makecell[l]{t3500芯\colorbox{lightgray}{320gb硬盘联想}\\\colorbox{lightgray}{b460el}本\colorbox{lightgray}{2699元}} \\
		\makecell[l]{Translation: \hl{2gb memory 320gb} \\\hl{hard disk Lenovo b460el} only \\\hl{2699 yuan}} & 
		\makecell[l]{Translation: gt core t3500\\ core \hl{Lenovo b460el}-tth \\only \hl{2699 yuan}} & 
		\makecell[l]{Translation: i3 core t3500 \\core \hl{Lenovo b460el}-tth \\only \hl{2699 yuan}} & 
		\makecell[l]{Translation: t3500 core \\\hl{320gb hard drive Lenovo}\\\hl{b460el} notebook \hl{2699 yuan}} \\
		\hline
		\makecell[l]{\colorbox{lightgray}{沪指}下挫报收\colorbox{lightgray}{3019.18点}\\\colorbox{lightgray}{创业板全线逆势飘红}} & 
		\makecell[l]{\colorbox{lightgray}{创业板逆势飘红沪指}跌\\1.23\%午后跌幅略有收缩} &
		\makecell[l]{\colorbox{lightgray}{沪}综\colorbox{lightgray}{指}最低跌至3012点\\午后跌幅略有收缩} & 
		\makecell[l]{\colorbox{lightgray}{沪}综\colorbox{lightgray}{指}报收\colorbox{lightgray}{3019.18点}\\\colorbox{lightgray}{创业板全线飘红}} \\
		\makecell[l]{Translation: \hl{Shanghai Composite} \\\hl{Index} fell to close at \hl{3019.18 points}\\ \hl{ChiNext} went \hl{red against the trend}\\\hl{across the board}} & 
		\makecell[l]{Translation: \hl{ChiNext} went \\\hl{red against the trend}, \\\hl{Shanghai index} fell 1.23\%, \\decline slightly contracted\\ in the afternoon} & 
		\makecell[l]{Translation: \hl{Shanghai} \\\hl{Composite Index} fell as low \\as 3012 points in the \\afternoon, decline narrowed\\ slightly} & 
		\makecell[l]{Translation: \hl{Shanghai} \\\hl{Composite Index} closed \\at \hl{3019.18 points}, \hl{ChiNext} \\was \hl{red across the board}} \\
		\hline
		\makecell[l]{\colorbox{lightgray}{报告}称\colorbox{lightgray}{2010年全球无线设备}\\\colorbox{lightgray}{收入}将达\colorbox{lightgray}{2355亿美元}} & 
		\makecell[l]{isuppli预计2011年\colorbox{lightgray}{全球无线}\\\colorbox{lightgray}{设备收入}将达2713亿美元} &
		\makecell[l]{isuppli预计\colorbox{lightgray}{全球无线设备收}\\\colorbox{lightgray}{入}到2011年将达2713亿美元} & 
		\makecell[l]{isuppli称\colorbox{lightgray}{2010年全球无线}\\\colorbox{lightgray}{设备收入}将达\colorbox{lightgray}{2355亿美元}} \\
		\makecell[l]{Translation: \hl{report} says \hl{global} \\\hl{wireless device revenue} to reach\\ \hl{\$235.5 billion} in \hl{2010}} & 
		\makecell[l]{Translation: isuppli expects \\\hl{global wireless equipment}\\ \hl{revenue} to reach \$271.3 \\billion in 2011} & 
		\makecell[l]{Translation: isuppli expects \\\hl{global wireless equipment}\\ \hl{revenue} to reach \$271.3 \\billion by 2011} & 
		\makecell[l]{Translation: isuppli says \\\hl{global wireless equipment}\\ \hl{revenue} will reach \hl{\$235.5}\\ \hl{billion} in \hl{2010}} \\
		\hline
		\makecell[l]{\colorbox{lightgray}{50城100楼盘}发放\colorbox{lightgray}{购房}\\\colorbox{lightgray}{消费券}\colorbox{lightgray}{购房者利益落空}} & 
		\makecell[l]{搜房网\colorbox{lightgray}{购房消费券}发行者\\全国各地媒体曝光} &
		\makecell[l]{房地产行业炒作沸沸扬扬\colorbox{lightgray}{消}\\\colorbox{lightgray}{费券}发行者是全国各地媒体} & 
		\makecell[l]{\colorbox{lightgray}{50}个\colorbox{lightgray}{城}市发券\colorbox{lightgray}{购房}\\\colorbox{lightgray}{消费券}覆盖\colorbox{lightgray}{100}多\colorbox{lightgray}{楼盘}} \\
		\makecell[l]{Translation: \hl{100 real estate} in\\ \hl{50 cities} issued \hl{consumer coupons}\\ \hl{interests of house buyers lost}} & 
		\makecell[l]{Translation: SouFun.com \\issuer of \hl{consumer coupons} \\is exposed by the media \\all over the country} & 
		\makecell[l]{Translation: real estate \\industry hyped, issuer of \\\hl{consumer coupons} is the media\\ from all over the country} & 
		\makecell[l]{Translation: \hl{50 cities} issued\\ \hl{consumer coupons} covering\\ more than \hl{100 real estate}} \\
		\hline
	\end{tabular}
	}
\caption{\label{casestudy1} Case study on generated headlines with Gigaword and THUCNews datasets. Gray parts are key information. The translation is supported by using Google Translate.}
\end{table*}
\end{CJK}

\begin{table}
	\centering\small
	\begin{tabular}{lcccc}
		\hline
		\multicolumn{1}{c}{\multirow{2}{*}{Method}} & \multicolumn{2}{c}{Gigaword-3k}    & \multicolumn{2}{c}{THUCNews-3k}     \\ 
		\cline{2-5} \multicolumn{1}{c}{} & Train & Test & Train & Test \\
		\hline
		Normal & 89s & 160s & 75s & 109s \\
		Normal+ & 90s & 149s & 72s & 101s \\
		Multi-task & 91s & 158s & 72s & 112s \\
		Duality & 496s & 167s & 376s & 115s \\
		\hline
	\end{tabular}
\caption{\label{cost} Time cost of model training for one epoch and inferring the testing sets with BART as the backbones.}
\end{table}

\paragraph{Performance on various sizes of datasets} To investigate more less-data situations, from the original large-scale corpora, we randomly collect different sizes of training datasets ranging from 1,000 (1K) to 10,000 (10K) with a interval of 1,000. Thus we have ten training sets for Gigaword and THUCNews respectively. Figure~\ref{datasize} illustrates the Rouge-1 and Micro-F1 scores correspondingly on language modeling metric and informative correctness on pre-trained BART. We can see the Duality and Normal+ methods can significantly improve the performance along with the increasing of data size, while Normal and Multi-task methods can obtain slight improvement. It is probably evident that leveraging the key information is beneficial for headline generation under less-data situations, and explicit modeling the information like Duality fine-tuning, instead of just putting key tokens ahead source document (i.e. Normal+), can capture more data knowledge especially when the dataset scale is small.

\subsection{Human Evaluation}

\paragraph{Human Grading} We perform human evaluation from the perspectives of readability and informativeness, which is to assess if the generated headlines are whether readable and informative for humans. We randomly sample 100 samples from the test sets of Gigaword and THUCNews datasets. We choose the generated headlines by using pre-trained BART models. Then the source documents, reference headlines, and generated headlines are randomly shuffled and shown to a group of people for evaluation. They cannot see the sources of headlines, i.e., from reference or inference. They need to judge the two aspects of readability and informativeness by giving an integer score in the range of 1-5, with 5 being perfect. Each sample is assessed by 5 people, and the average scores are used as the final score. To keep the labeling quality and further reduce bias, we normalize the scores of each people by z-score normal distribution. 

As shown in Table~\ref{human}, we find that the Duality gets best or best -comparable readability scores among the three evaluated methods. 
For the informativeness, Duality method can significantly perform best, which demonstrates its effectiveness to generate informative headlines. Comparing the scores of generated headlines and ground-truth references, there is still a large gap between model-generated and human-composed headlines, especially on the Chinese dataset THUCNews. 

\paragraph{Case Study} We analyze 50 test samples from the Gigaword and THUCNews, and compare the generated headlines with different methods. Table~\ref{casestudy1} shows the results of respective five samples. The ground-truth or generated key information are marked by gray highlights. We find that Duality performs better than other methods in most cases. For example, in the second and fifth cases of Gigaword cases in Table~\ref{casestudy1}, Duality can generate more key information tokens than others, as well as the examples from THUCNews cases. We also observe that Dulity could perform better on Chinese data, perhaps because Chinese headlines have higher ratio of key tokens among the token sequence.

\paragraph{Error Analysis} From the above 50 test samples, we also observe some bad cases generated by our method. We categorize them to several common types of error: incomplete key information (8 cases), repeats (5 cases), wrong key information (4 cases), and not coherent language (8 cases). And they should be investigated in the future work.

\subsection{Computational Cost Analysis}

During the model training phase, since Duality fine-tuning method should learn two separate dual models for each task, i.e. one more than the other baselines, it is inevitable that Duality method would spend more computing time and twice memory space. During the testing phase, since we only use one model to generate headlines, the computing cost of Duality method is comparable to the others. Table~\ref{cost} shows the computing time cost of each method with BART as pre-trained models on 3k training datasets and 500 testing datasets via one 32G-V100 GPU. We can see that although training one-epoch dual models would spend more time than other methods, the absolute spent time is still acceptable and efficient considering the less-data situations and the performance improvement.

\section{Conclusion}

In this paper, we introduce a novel task that how to improve the performance of less-data constrained headline generation. We highlight to explicitly exploit the key information, and propose a novel duality fine-tuning method which firstly integrates dual learning paradigm and fine-tuning paradigm for less-data generation. The proposed method should obey the probabilistic duality constraints, which are critical to model multiple tasks. Therefore, the method can model more supervised information, learn more knowledge, and train more powerful generative models. Our method can also be generally applied to both autoregressive and encoder-decoder generative regimes. We collect various sizes of small-scale training datasets from two public corpora in English and Chinese, and the extensive experimental results prove our method effectively improve the readability and informativeness of generated headlines with different pre-trained models.



\section*{Acknowledgements}
We thank all the anonymous reviewers for their valuable feedback and insightful comments.

\bibliography{anthology,custom}
\bibliographystyle{acl_natbib}



\end{document}